# CORI: CJKV Benchmark with Romanization Integration - A step towards Cross-lingual Transfer Beyond Textual Scripts

**Hoang H. Nguyen**[1], **Chenwei Zhang**[2], **Ye Liu**[3], **Natalie Parde**[1], **Eugene Rohrbaugh**[4], **Philip S. Yu**[1]

[1] Department of Computer Science, University of Illinois at Chicago, Chicago, IL, USA
[2] Amazon, Seattle, WA, USA [3] Salesforce Research, Palo Alto, CA, USA
[4] Harrisburg University of Science and Technology , Harrisburg, PA, USA
{hnguy7,parde,psyu}@uic.edu, cwzhang@amazon.com
yeliu@salesforce.com, gene.rohrbaugh@gmail.com

**Abstract**

Naively assuming English as a source language may hinder cross-lingual transfer for many languages by failing to consider the importance of language contact. Some languages are more well-connected than others, and target languages can benefit from transferring from closely related languages; for many languages, the set of closely related languages does not include English. In this work, we study the impact of source language for cross-lingual transfer, demonstrating the importance of selecting source languages that have high contact with the target language. We also construct a novel benchmark dataset for close contact Chinese-Japanese-Korean-Vietnamese (CJKV) languages to further encourage in-depth studies of language contact. To comprehensively capture contact between these languages, we propose to integrate Romanized transcription beyond textual scripts via Contrastive Learning objectives, leading to enhanced cross-lingual representations and effective zero-shot cross-lingual transfer.

## 1. Introduction

Language is fluid and ever-evolving, with specific languages often borrowing terms and even grammatical structure from one another as a byproduct of voluntary or forced interaction. *Language contact* defines the emerging connections among languages throughout the course of history, geography, and social development (Matras, 2020; Mufwene and Escobar, 2022). For example, modern English (EN) is a Germanic language, yet it contains many Latinate and Greek terms that entered the vernacular via religious missionaries or war and conquest, as well as other terms that later entered the lexicon via colonization (van Gelderen, 2014). Likewise, Japanese (JA), Korean (KO) and Vietnamese (VI) share a large number of cognates with Chinese (ZH) due to influence through colonization, trading throughout history (Li, 2019; Hickey, 2020). This influence is best reflected through the Kanji characters in JA writing system, which significantly overlaps with ZH character writing systems. As observed in Table 1, both ZH and JA writing systems utilize 文学 and 学者 to denote words *scholar* and *classical* respectively.

Despite recent advances in the study of cross-lingual transfer, most cross-lingual works to date have restricted themselves to two major assumptions: (1) English (EN) is considered the single source language, despite its oftentimes limited or absent contact with the target language(s) (Muller et al., 2021; Fujinuma et al., 2022); and (2) cross-lingual transfer remains restricted to orthographic language contact (Nguyen and Rohrbaugh, 2019; Yang et al., 2022). As language acquisition naturally benefits from various linguistic modalities, including the textual writing scripts and aural signals, which can be conveyed through orthographic and phonemic transcription respectively (Volterra and Iverson; Goswami, 2022), the aforementioned constraint hinders progress in the cross-lingual field, by minimizing the amount of relevant information that models may be able to access (Fujinuma et al., 2022; Nguyen et al., 2023c).

Because of language contact, there are many lexical items shared among CJKV languages. However, these similarities may be obscured by the different writing systems used across the languages. Borrowed words that may sound very similar in spoken form may nevertheless be represented in the writing systems with completely different symbols. As observed in Table 1, the orthographic term 古典 in ZH shares little similarity with VI counterpart. However, the corresponding phonemic transcription represented via Romanization (gǔdiǎn) is more similar to the corresponding VI (Cổ điển). The same phenomenon can also be observed in JA-KO pair (古典 (koten) vs 고전 (gojeon)) and VI-KO pair (Khoa học (Khoa học) vs 과학 (gwahak)). Moreover, Table 1 reveals the shared properties of CJKV languages in which meaningful words are formed by combining surrounding tokens. For instance, four characters 形而上学 are combined together to form the meaningful word “Metaphysical”. The same behavior can be observed in the corresponding VI (Siêu hình) and KO (형이상학). Although the current multiilngual benchmark dataset such as XTREME (Hu et al., 2020) offers extensive language coverage, the aforementioned shared properties among CJKV are often overlooked during preprocessing.

We address these limitations through the following contributions:

- We study the impact of source language selection in zero-shot cross-lingual transfer among languages with historical language contact.
- We construct a comprehensive set of language understanding (NLU) tasks for CJKV languages that (1) better presents the shared characteristics of CJKV languages than the existing multilingual XTREME benchmark, (2) provides additional phonemic signals captured through Romanized transcriptions beyond orthographic textual scripts.
- We integrate phonemic information from Romanization with Orthographic texts via Contrastive Learning to enhance textual representation across CJKV languages, leading to improved cross-lingual representation and downstream task performance.[1]

## 2. Related Work

**Cross-lingual Transfer.** Recent works on cross-lingual transfer have focused on generating multilingual contextualized representations for different languages via Pre-trained Language Models (PLMs) trained using multilingual dictionaries (Qin et al., 2021; Winata et al., 2023) and/or machine translation approaches (Fang et al., 2021; Liu et al., 2021c; Reid and Artetxe, 2023). Qin et al. (2021) proposed a comprehensive code-switching technique via random selection of languages, sentences, and tokens to enhance multilingual representations, leading to improved downstream task performance on target languages. On the other hand, other approaches have leveraged machine-translated parallel corpora to (1) distill knowledge from source languages to target languages (Fang et al., 2021) or (2) augment source language data with target language knowledge during training (Yang et al., 2022; Zheng et al., 2021).

However, current cross-lingual efforts concentrate on transferring from a single source language (EN) to multiple target languages (Hu et al., 2020; Yang et al., 2022). Under parameter capacity constraints, cross-lingual transfer has been shown to be biased towards high-resourced languages which share similar scripts and possess larger corpora of unlabeled data during pre-training (Fujinuma et al., 2022). Recent works (Adams et al., 2017; Nguyen et al., 2023c) attempt to bridge the gap between languages across different scripts by introducing phonemic transcription. However, the introduction of the International Phonetic Alphabet (IPA) to this process requires additional pre-training objectives to effectively unify the two modalities. Unlike previous works, we leverage natural Romanized transcriptions to capture phonemic properties.

**Contrastive Learning.** Contrastive Learning (CL) has been widely leveraged as an effective representation learning mechanism (Oord et al., 2018; Chen et al., 2020). The goal of CL is to learn the discriminative features of instances via different augmentation methods. In Natural Language Processing (NLP), CL has been adopted in various contexts ranging from text classification (Wei and Zou, 2019) and representation learning (Gao et al., 2021; Nguyen et al., 2023b) to question answering (Xiong et al., 2020; Liu et al., 2021b). Unlike previous works, we leverage two language modalities (orthographic and Romanized transcriptions) to generate multi-view augmentations as effective mechanisms to improve cross-lingual representations.

## 3. Problem Formulation

We study the problem of cross-lingual transfer in a multilingual setting where there exists annotated data $D$ collected from a high-resource language $S$, such that $D_S^{train} = \{(X_i^{(S)}, Y_i^{(S)})\}_{i=1}^{N_s}$ where $N_s$ denotes the sample size of the source language training data. Likewise, there exists unlabeled data collected from a set of low-resource target languages $T = \{T_1, ..., T_n\}$, denoted as $D_{T_i}{}^{test} = \{(X_j^{(T)})\}_{j=1}^{N_{t_i}}$ where $N_{t_i}$ denotes the sample size of the $i$-th target language inference dataset. Depending on the downstream tasks, $Y_i$ can be either a single utterance label (Sentence-level task), a sequence of token labels (Token-level Task), or extracted span labels (Question Answering (QA)).

Our work encompasses a comprehensive set of Natural Language Understanding (NLU) tasks among CJKV languages as briefly summarized below. For further details of individual task objectives and metrics, we refer readers to (Hu et al., 2020).

**Sentence-level Task (PAWSX, XNLI)** concentrates on overall sentence-level semantic understanding of the given input paired utterance. The two covered tasks are multilingual Paraphrase Identification (PAWSX) and Natural Language Inference (XNLI).

**Token-level Task (UDPOS, PANX)** requires deeper understanding on token-level which re-quires additional grammatical structural knowledge beyond overall semantics. Representative tasks include Part-of-speech (POS) Tagging through UDPOS dataset and Named Entity Recognition (NER) via PANX dataset.

[1]Our code and datasets are publicly available at https://github.com/nhhoang96/benchmark_cjkv

Table 1: Orthographic and Romanized representations (abbreviated as Ortho and Roman) of a sample sentence across CJKV languages where **colored segments** denote the corresponding semantic **words** defined in Section 5. (.) denotes the specific name of Romanization system of the respective language. The first sentence for each language denotes the currently preprocessed XTREME benchmark dataset.

| Language | Input Type | Sample input sentence |
|---|---|---|
| EN | Ortho | He was a scholar in Metaphysical Literature, Theology and Classical sciences. |
| ZH (source) | Ortho | 他是形而上学文学、神学和古典科学方面的学者。 |
| | Ortho (seg) | 他 // 是 // 形而上学 // 文学 // 、 // 神学 // 和 // 古典 // 科学 // 方面 // 的 // 学者 // 。 |
| | Roman (Pinyin) | tā // shì // xíngérshàngxué // wénxué // 、 // shénxué // hé // gŭdiăn // kēxué // fāngmiàn // de // xuézhě // 。 |
| VI (target) | Ortho | Ông là một học giả về Văn học Siêu hình, Thần học và Khoa học Cổ điển. |
| | Ortho (seg) | Ông // là // một // học giả // về // Văn học // Siêu hình // , // Thần học // và // Khoa học // Cổ điển // . |
| | Roman | Ông // là // một // học giả // về // Văn học // Siêu hình // , // Thần học // và // Khoa học // Cổ điển // . |
| JA (target) | Ortho | 彼は形而上学文学、神学、古典科学の学者でした。 |
| | Ortho (seg) | 彼 // は // 形而上学 // 文学 //、 神学 、 // 古典 // 科学 //の// 学者 //でし // た // 。 |
| | Roman (Romaji) | kare // ha // keiji ue gaku // bungaku // , // shingaku // , // koten // kagaku // no // gakusha // deshi // ta // . |
| KO (target) | Ortho | 그는형이상학문학, 신학및고전과학의학자이었습니다. |
| | Ortho (seg) | 그 // 는 // 형이상학 // 문학 //, // 신학 // 및 // 고전 // 과학 // 의학자 // 이 // 었 // 습니다 // . |
| | Roman (Romaja) | eu // neun // hyeongisanghak // munhak // , // sinhak // mit // gojeon // gwahak // uihakja // i // eot // seupnida // . |

Table 2: Preliminary study on the impact of source language (ZH) on target JKV languages. Standard deviation denotes the performance variation across 3 target JKV languages. Performance of individual JKV language is reported in Table 7 in the Appendix.

| **Source** | **Sentence-level** | | **Token-level** | | **Question Answering** | |
|---|---|---|---|---|---|---|
| | **PAWSX** | **XNLI** | **UDPOS** | **PANX** | **XQuAD** | **MLQA** |
| Metric | Acc | Acc | F1 | F1 | EM | EM |
| EN | 73.53 $\pm$ 6.86 | 68.35 $\pm$ 2.05 | 40.09 $\pm$ 10.32 | 30.36 $\pm$ 19.09 | 28.38 $\pm$ 18.00 | 22.11 $\pm$ 8.76 |
| **ZH** | **79.61** $\pm$ **5.24** | **70.39** $\pm$ **1.60** | **47.82** $\pm$ **1.98** | **37.29** $\pm$ **9.18** | **33.95** $\pm$ **14.73** | **29.95** $\pm$ **5.88** |

**Question Answering Task (XQuAD, MLQA)** extracts the answer spans existent in the given Context when provided with input corresponding Question. The task assumes the answers are existent in the given Context; therefore, within the scope of our study, any samples without existent answers in the contexts are removed from both training and evaluation. Unlike (Hu et al., 2020), we only report Exact Match (EM) for QA task since it is considered a stricter evaluation measure and requires models to extract truly meaningful answers.

## 4. Preliminary Study

We conduct a preliminary study to examine the effects of source language selection on target languages. In our case, we compare using EN against ZH as the source language to transfer knowledge towards other languages among the CJKV group. ZH is considered the common source language for CJKV due to its historical longevity and significant impact on the development of the remaining languages (Li, 2019; Hickey, 2020). We fine-tune a backbone XLM-R-base PLM (Conneau et al., 2020) with downstream task objectives for each source-target JKV language pair and report average performance together with standard deviation across the three target languages JA, KO, VI.

Overall, based on our preliminary study reported in Table 2, we observe the following:

- Average zero-shot cross-lingual transfer performance for target languages across all tasks increases significantly when ZH is leveraged as the source language. On UDPOS task, the average performance of JKV languages reaches F1 of 47.82 as compared to 40.09 when EN is considered source language.

- When ZH is leveraged as source language instead of EN, the performance standard deviations across target JKV languages diminished (19.09 for EN source vs 9.18 when ZH source on PANX task), implying that appropriately selected source languages can reduce the performance gap across target languages.

Our observations are consistent with prior study (Fujinuma et al., 2022; Yang et al., 2022) demonstrating that PLMs when trained on only EN can be biased towards Latin-based languages, resulting in poor performance for non-Latin based

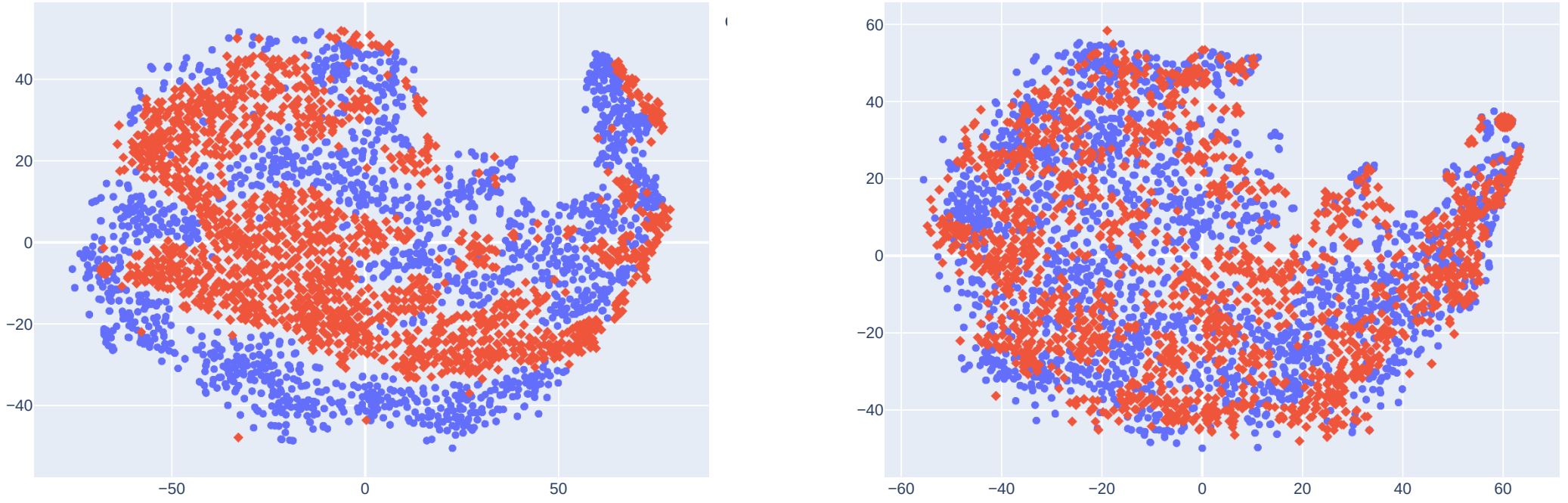

(a) EN source (blue), JA target (red)　　(b) ZH source (blue), JA target (red)

Figure 1: Representation visualization of parallel sentences between source language and target language (EN-JA (left) and ZH-JA (right)) of the fine-tuned XLM-R model on parallel PAWSX test set. The significant overlapping representation of ZH-JA demonstrates the lower representation discrepancy and higher language contact between source and target language.

Table 3: Average CKA Score (Kornblith et al., 2019) Comparison between selected source languages (EN vs ZH) across 3 target JKV languages on parallel PAWSX test set. **O,R** denote **Orthographic** and **Romanized** transcription input respectively. The higher CKA score demonstrates the lower representation discrepancy and implies the higher language contact.

| **Source** | **EN (O)** | **EN (O+R)** | **ZH (O)** | **ZH (O+R)** |
|---|---|---|---|---|
| JA | 0.0095 | 0.0638 | 0.0830 | 0.1428 |
| KO | 0.0175 | 0.0571 | 0.0120 | 0.1184 |
| VI | 0.0407 | 0.0572 | 0.0270 | 0.1200 |
| **Average** | **0.0225** | **0.0594** | **0.0403** | **0.1271** |

languages and significant gaps in performance between Latin-based and non-Latin-based languages as demonstrated in higher performance standard deviations (10.32 on UDPOS for EN) when compared to ZH (1.98 on UDPOS for ZH).

Figure 1 further demonstrates the representation discrepancy between EN-JA in comparison with ZH-JA. Following Conneau and Lample (2019); Yang et al. (2022), we adopt the Centered Kernel Alignment (CKA) (Kornblith et al., 2019) as a quantitative measure of representation discrepancy among evaluated languages. The significant higher CKA score for ZH-JA representation pair (0.0830) as compared to EN-JA pair (0.0095) presented in Table 3 validates the representation gap existent from model trained purely on EN, prompting for careful consideration of source language in zero-shot cross-lingual transfer. Based on Table 3, we also observe the average CKA score between ZH and target JKV languages is significantly higher than the EN source (0.0403>0.0255). As Romanization is introduced in conjunction with orthographic transcription as further detailed in Section 5.3 and 6, the average CKA score between ZH and target JKV languages further improves and widens the gap with EN counterpart (0.1271>0.0594). These observations confirm our intuition that ZH serves as a more representative source language for CJKV languages when compared to EN and the selection of ZH as source language can further enhance downstream task performance on target JKV languages. These benefits underscore the necessity of a dataset tailored specifically for CJKV.

## 5. Dataset Construction

XTREME Dataset (Hu et al., 2020) provides a comprehensive multilingual benchmark for multiple NLU tasks across languages from diverse typology. However, when it comes to the study of CJKV languages on the existing XTREME, there exist three major challenges: language availability, pre-segmentation, and orthographic contact limitation. We elaborate on each of those below.

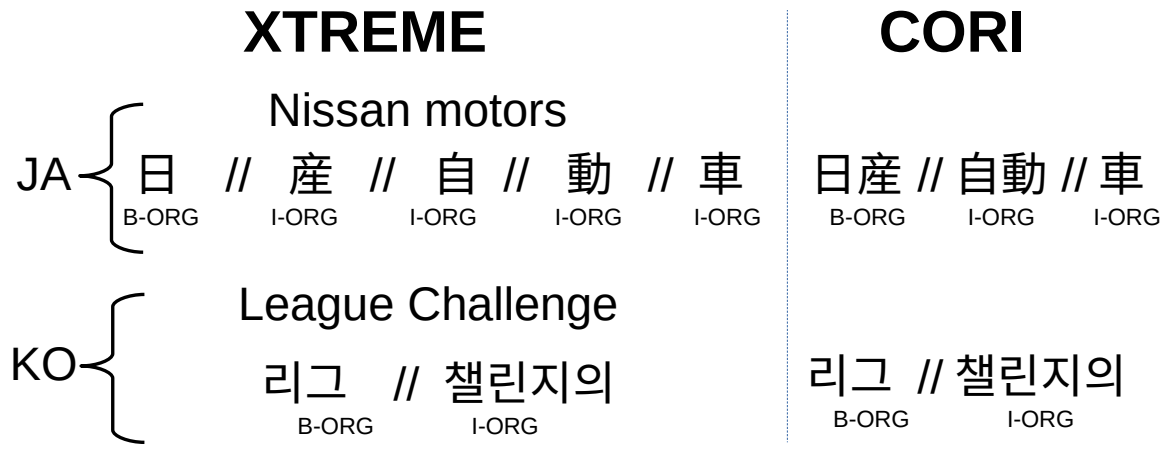

Figure 2: Example of label inconsistency between JA and KO languages on PANX dataset of XTREME and the alleviation provided in CORI.

Table 4: Details of CORI benchmark dataset. (✗) and (✓) denote unavailable and available data existent in XTREME benchmark respectively . MT, SEG, ROM correspond to Machine Translation, Pre-segmentation, Romanization processing respectively as described in Section 5. Y, N denote if the corresponding preprocessing is conducted for the dataset or not.

| | | MT | SEG | ROM | ZH | | JA | KO | VI |
|---|---|---|---|---|---|---|---|---|---|
| | | | | | Train | Dev | Test | Test | Test |
| **Sent-level** | PAWSX | Y | Y | Y | 49.4k (✗) | 2k (✓) | 2k (✗) | 2k (✓) | 2k (✗) |
| | XNLI | Y | Y | Y | 392.7k (✓) | 2.49k (✓) | 5.01k (✗) | 5.01k (✗) | 5.01k (✓) |
| **Token-level** | UDPOS | N | N | Y | 10k (✓) | 1.6k (✓) | 2.5k (✓) | 4.7k(✓) | 0.8k (✓) |
| | PANX | N | Y | Y | 20k (✓) | 10k (✓) | 10k (✓) | 10k (✓) | 10k (✓) |
| **Question Answering** | XQuAD | Y | Y | Y | 80.1k (✗) | 8.87k (✗) | 1.19k (✗) | 1.19k (✗) | 1.19k (✓) |
| | MLQA | Y | Y | Y | 80.1k (✗) | 8.87k (✗) | 11.24 k (✗) | 11.24k (✗) | 11.24k (✓) |

**Language Availability.** Despite the multilingual coverage of XTREME, the language support for each task is not equal. For instance, while JA has train, dev, and test sets for token-level part-of-speech (POS) tasks, there exists no JA test set for sentence-level PAWSX task evaluation. Therefore, the evaluation of existing works does not fully facilitate deep understanding of source and target language connections.

**Pre-segmentation.** The important shared characteristics of segmentations among CJKV languages are inherently ignored and can be lost during multilingual pre-processing and translations. For consistency purposes, in our work, we define the terms **word** and **token** across CJKV languages and cohere to these definitions throughout the manuscript. For CJK, we define each character as a token and the combination of tokens after segmentation as a word. For VI, we refer to whitespace-separated terms as tokens and the combination of tokens after segmentation as a word. Each word carries a complete semantic meaning and can be formed by a single token or multiple tokens.

For CJK, there exist no consistent rules of using whitespace delimiters to differentiate tokens from words. On the other hand, VI is written in Latin script but still whitespace does not separate meaningful units of the sentence as it does in EN. For all of our languages of interest (CJKV), instead meaningful word units are formed by combining individual tokens (Choi et al., 2009; Nguyen and Nguyen, 2020; Sun et al., 2021).

Additionally, as entries from bilingual dictionaries are stored in terms of semantic words, pre-segmented words allow for efficient lookups when generating code-switched language. The issue of pre-segmentation leads to inconsistent annotations across languages. Specifically, within the XTREME benchmark, on PANX datasets for NER task, annotations for KO are generated at the word level (리그 meaning League) while JA labels are generated at the token level (日 as a part of 日産 word denoting Nissan meaning), as demonstrated in Figure 2. Through additional pre-segmentation procedures, our CORI dataset provides more consistent word-level annotations across CJKV.

**Orthographic Contact Limitation.** Prior works assume language contact is limited to orthographic textual scripts (Zheng et al., 2021; Yang et al., 2022). However, languages can interact with one another via other language modalities. In fact, as observed in Table 1, while ZH and JA share orthographic contacts (shared 形而上学 word in textual scripts (meaning Metaphysical)), ZH and VI contain subtle phonemic contacts captured via Romanization such as gǔdiǎn (ZH) vs Cổ điển (VI), denoting "Classical"(EN) meaning.

Our presented CJKV Benchmark with Romanization Integration, namely CORI, aims to address the aforementioned shortcomings and allows for in-depth study on the connections of CJKV languages across multi-level understanding tasks. CORI contains 6 sub-datasets across different NLU tasks mentioned in Section 3 in which training and validation sets for ZH (source language) and testing set for 3 target JKV languages are provided. Additional detailed summary statistics regarding this dataset are provided in Table 4.

## 5.1. Language Availability

As token-level data for CJKV languages are all provided in XTREME benchmark datasets, we focus on generating translations for sentence-level and question answering tasks. Since these two tasks do not suffer dramatically from label projection challenges as compared to token-level tasks (Agić et al., 2016), we adopt the off-the-shelf Google API Machine Translation (MT) Tool[2] to generate translations for target CJKV languages. The objective is two-fold (1) MT tool provides an automatic and human-free approach that can easily scale and be applied towards other datasets to generate CJKV equivalents, (2) Off-the-shelf MT tools decouple MT errors from cross-lingual transfer learning, allowing for direct evaluation of the studied cross-lingual methods.

[2]https://pypi.org/project/googletrans-py/

**Sentence-Level Task.** As both PAWSX and XNLI require paired inputs, for each entry we simply translate each input separately while preserving the ground truth labels for the pairs.

**Question Answering Task.** We adopt a template-based machine translation approach (Liu et al., 2021a) to generate QA datasets for CJKV languages. Specifically, given individual *Context*, *Answer* and *Question* triplets from the source EN input, we mask out the ground truth *Answer* within the *Context* using a special token. The masked *Context* and the ground truth *Answer* are then translated into the target languages separately. Eventually, we replace the special mask token with the translated *Answer* to construct the translated *Context*. As *Question* does not involve the ground truth *Answer*, we translate it independently similarly to the process used for sentence-level task preprocessing.

## 5.2. Pre-segmentation

Generating pre-segmentation on the word-level semantics requires in-deptth knowledge of specific target languages. For this reason, we leverage language-specific segmentation tools to provide an automatic alternative to human-dependent approaches. Despite imperfections, produced segmentations from these tools provide transparent word-level semantic boundaries for multilingual PLM tokenizer (Conneau et al., 2020), which facilitates the vocabulary knowledge acquisition for individual target languages (Nakata and Suzuki, 2019; Choi et al., 2021). Moreover, pre-segmentation allows for effective aggregation of Orthographic and Romanized transcription inputs that are differently tokenized by PLMs. We leverage pykakasi[3], konlpy[4], jieba[5] and pyvi[6] for JA, KO, ZH, VI for pre-segmentation process respectively. As UDPOS dataset has already been pre-segmented, we skip processing for this task as displayed in Table 5.

## 5.3. Articulatory Signals

Language contact is not necessarily restricted to orthographic textual scripts, especially in cases when most contact occurred before the present writing system was established or popularized; a textbook example of this would be Hindi and Urdu, which are highly mutually intelligible but use completely different orthographic scripts (Ahmad, 2011). Thus, it is also beneficial to capture linguistic connections via other modalities of languages. For instance, the orthographic representation of the term “classical” in ZH (古典) and VI share little similarity with one another. However, on the phonemic level represented via Romanized transcription, gǔdiǎn in ZH is more similar to the corresponding VI (Cổ điển).

Articulatory signals can be extracted from multiple potential sources such as audio wave signals, IPA transcription, or simple Romanization. While audio signals can capture a wide range of aspects regarding speech, they also contain additional noise from human accents and background noise (Wang et al., 2022; Hu et al., 2023). IPA provides a unified system for phonemic transcription, but it is unnatural and generally unsupported for CJKV languages (Mortensen et al., 2016) and necessitates additional pre-training for PLMs due to the introduction of additional special tokens (Nguyen et al., 2023c). Given the shortcomings of audio signals and IPA for capturing articulatory signals, we proceed with a simple Romanization that captures articulatory signals while maintaining the simplicity of phonemic transcription.

To Romanize text, we leverage language-specific Romanization tools nagisa[7], korean_romanizer[8], pypinyin[9] , resulting in Hepburn Romanization for JA, and Romaji for KO, Pinyin for ZH respectively. As VI is already written in Latin script, we consider the original orthographic scripts as its own Romanized version. The sample outcomes of both Pre-segmentation and Romanization are displayed in Table 1.

# 6. Proposed Framework

In this section, we introduce a simple proposed framework to integrate textual scripts with Romanized transcriptions via Contrastive Learning (CL) objectives as depicted in Figure 3.

**Phonemic-Orthographic Integration** Depending on the tokenizer of PLM, Romanized and Orthographic transcriptions might be tokenized into unequal number of sub-tokens. Therefore, we align the semantics of these corresponding representations on the preprocessed word-level semantics introduced in Section 5. Formally, given an $i$-th input source language utterance with the length of $M$ orthographic tokens $x_i^{(S)} = [x_{i,1}^{(S)}, x_{i,2}^{(S)} ..., x_{i,M}^{(S)}]$ and the corresponding phonemic transcriptions $z_i^{(S)} = [z_{i,1}^{(S)}, z_{i,2}^{(S)} ..., z_{i,M}^{(S)}]$:

$$I_i^{(S)} = [F(x_S), F(z_S)] \quad (1)$$

where $F(\cdot)$ denotes the Transformer Encoder layers of PLM and $[\cdot]$ represents function to unify the two representations produced by the Transformers. As our major objective is to verify the benefits of joint representation, we leverage a simple concatenation operation.

[3] https://pypi.org/project/pykakasi/
[4] https://konlpy.org/en/latest/
[5] https://pypi.org/project/jieba/
[6] https://pypi.org/project/pyvi/
[7] https://pypi.org/project/nagisa/
[8] https://github.com/osori/korean-romanizer
[9] https://pypi.org/project/pypinyin/

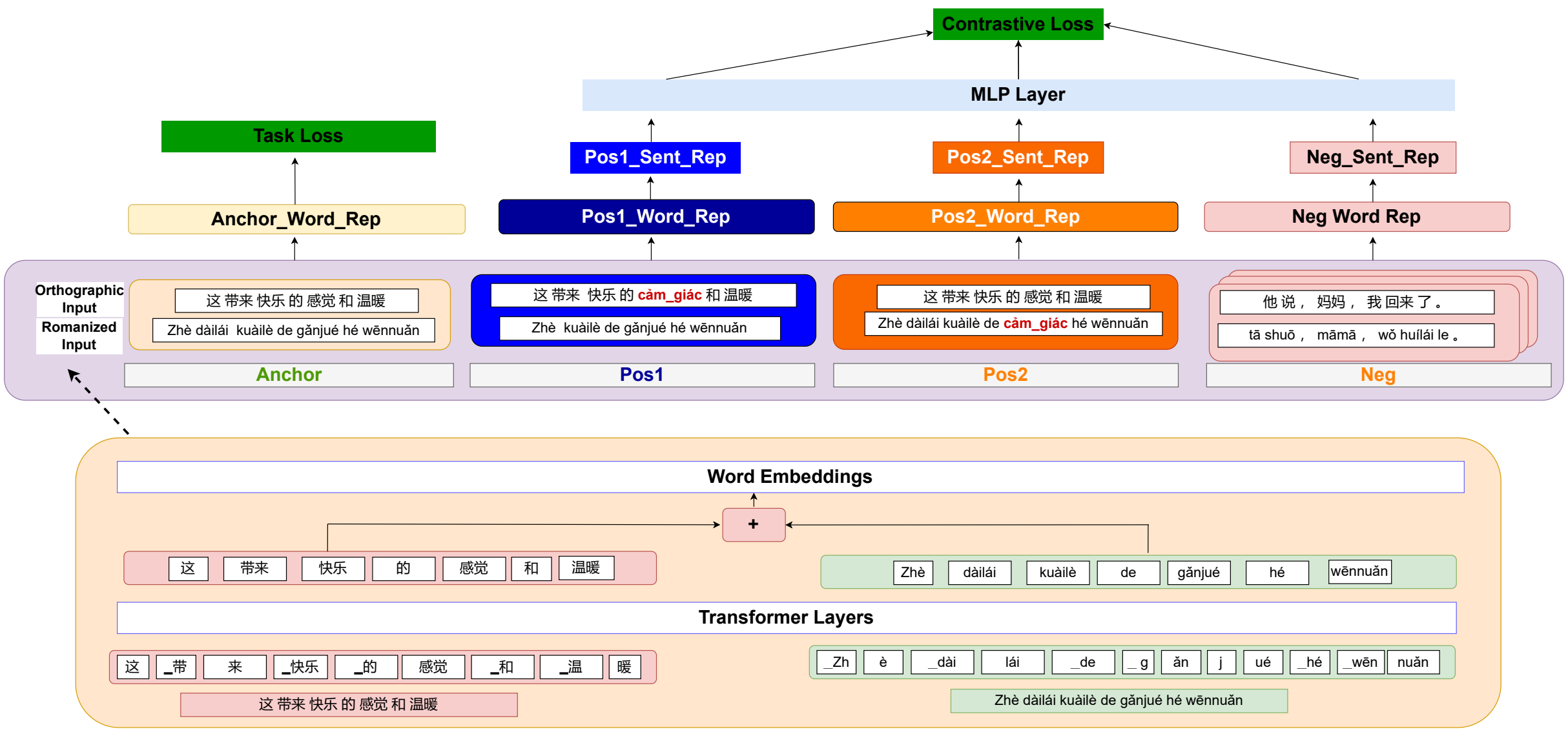

Figure 3: Illustration of the Proposed Model Overview. The sample utterance means "This brought a feeling of joy and warmth" in EN. *"感觉"* and *"cảm_giác"* (equivalent to *"feeling"* in EN) are semantically similar words in multilingual dictionaries. **Pos1, Pos2** represent two augmented views of the original **Anchor** sample via code-switching on Orthographic and Romanized transcriptions respectively. **Neg** denotes in-batch negative samples similarly defined in (Chen et al., 2020).

**Multi-view Augmentation** An essential component of CL is data augmentation. Previously, due to constraints from the single orthographic representation, augmentations are restricted to textual scripts. Taking advantage of the existence of parallel Romanized and Orthographic transcriptions, we propose a simple and parallel code-switching augmentations on each linguistic modality separately to generate multi-view augmentations of the language input as presented in Equation 2 and 3.

$$\tilde{I}_{i,1} = [F(CS(x_i)), F(z_i)] \\ \tilde{I}_{i,2} = [F(x_i), F(CS(z_i))] \tag{2}$$

where $CS(.)$ denotes code-switching operations on the corresponding input. We adopt the similar code-switching procedures proposed in (Qin et al., 2021) under multilingual settings with ratio of 0.5. Then, we apply the average pooling operation to obtain the sentence-level representation of each augmented view. Following (Chen et al., 2020), we define the Multi-layer Perceptron (MLP) as projection head ($g(\cdot)$) while our Transformer Layers ($F(\cdot)$) represents the encoder function:

$$\tilde{I}_1 = g(MeanPooling([\tilde{I}_{i,1}, ...., \tilde{I}_{i,M}])) \\ \tilde{I}_2 = g(MeanPooling([\tilde{I}_{i,2}, ...., \tilde{I}_{i,M}])) \tag{3}$$

CL objective is leveraged to encourage agreement of different augmented views of the original anchor input as formulated in Equation 4:

$$\mathcal{L}_{cl} = -log \frac{\exp(cos(\tilde{I}_1, \tilde{I}_2)/\tau)}{\exp(cos(\tilde{I}_1, \tilde{I}_2)/\tau) + \exp(\tilde{I}_1, I_{-})/\tau} \tag{4}$$

where $\tau$ denotes the soft temperature hyperparameter and $I_{-}$ denote in-batch negative samples.

**Task Objective** Depending on the downstream task, the exact output labels and learning objectives can be different. Without loss of generality, we define $Y^j$ as the downstream task label for j-th sample in the source language. Therefore, the task objective can be defined as:

$$\mathcal{L}_{task} = \frac{1}{N_s} \sum_{j=1}^{N_s} CrossEntropy(\delta([I^j_{i,1}, ...I^j_{i,M}]), Y^j) \tag{5}$$

where $\delta(\cdot)$ is the projection layer for target downstream tasks. The overall learning objective ($\mathcal{L}$) is summarized as: $\mathcal{L} = \mathcal{L}_{task} + \mathcal{L}_{cl}$.

## 7. Experiment

**Benchmark Baseline** We compare our proposed method with the current relevant state-of-the-art multilingual approaches under zero-shot cross-lingual transfer settings. Parallel corpora created using machine translation are expensive to obtain, prone to errors across languages from diverse typology, and require language-specific pairs for individual tasks. Therefore, we assume that translations are unavailable under zero-shot cross-lingual transfer settings. For baselines leveraging machine translation data under zero-shot

Table 5: Experimental results across multiple-level NLU tasks on XTREME and CORI benchmark **test** datasets with standard deviations across 3 target JKV languages when trained on source language ZH. Reported performance for each target language is averaged over 3 runs. **Bold** and † denote the best and second-to-best reported average performance respectively for each task/ column.

| Model | Sentence-level | | Token-level | | Question Answering | |
|---|---|---|---|---|---|---|
| | PAWSX | XNLI | UDPOS | PANX | XQuAD | MLQA |
| | Acc | Acc | F1 | F1 | EM | EM |
| XLM-R (raw XTREME) | 76.13 ± 10.59 | 59.18 ± 19.89 | – | 26.89 ± 10.29 | 30.31 ± 17.18 | 25.87 ± 14.18 |
| XLM-R | 79.61 ± 5.24 | 70.39 ± 1.60 | 47.82 ± 1.98 | 37.29 ± 9.18 | 33.95 ± 14.73 | 29.95 ± 5.88 |
| **Ours** | 81.38 ± 4.48 † | 71.45 ± 2.55 † | **51.75 ± 1.49** | **44.01 ± 12.96** | 39.77 ± 10.50 † | **35.83 ± 5.92** |
| CoSDA-ML | – | – | 50.48 ± 3.60 | 29.73 ± 13.91 | – | – |
| FILTER | 79.66 ± 5.78 | 68.06 ± 1.09 | 42.27 ± 17.78 | 29.34 ± 4.62 | 32.10 ± 10.14 | 28.06 ± 6.89 |
| xTune | **81.72 ± 4.91** | **72.25 ± 1.31** | 50.75 ± 1.16 | 40.53 ± 18.17 | **42.07 ± 11.25** | 33.13 ± 10.06 † |
| X-MIXUP | 80.41 ± 6.43 | 71.20 ± 2.98 | 50.84 ± 0.82 † | 40.71 ± 3.62 † | 32.69 ± 17.10 | 28.24 ± 7.46 |
| PhoneXL | – | – | 49.45 ± 1.77 | 38.26 ± 17.19 | – | – |

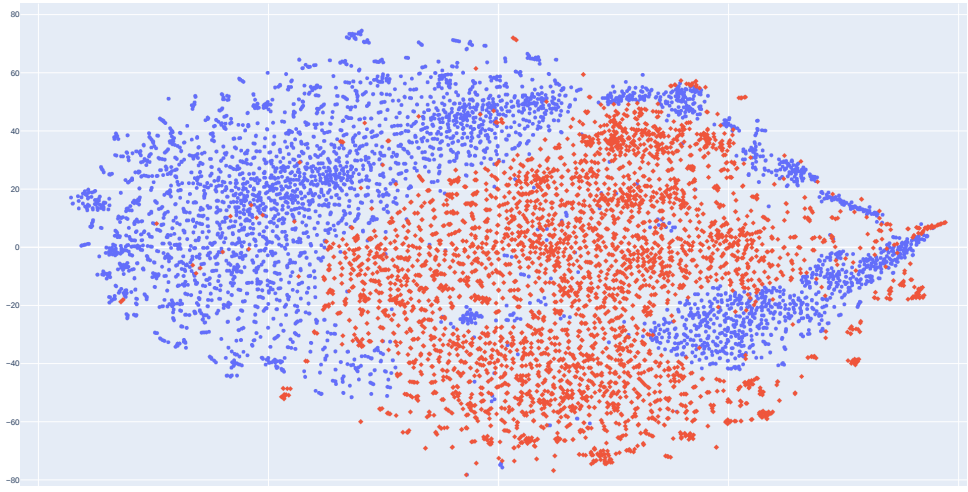

(a) Without Romanized transcriptions (CKA Score=0.0012)

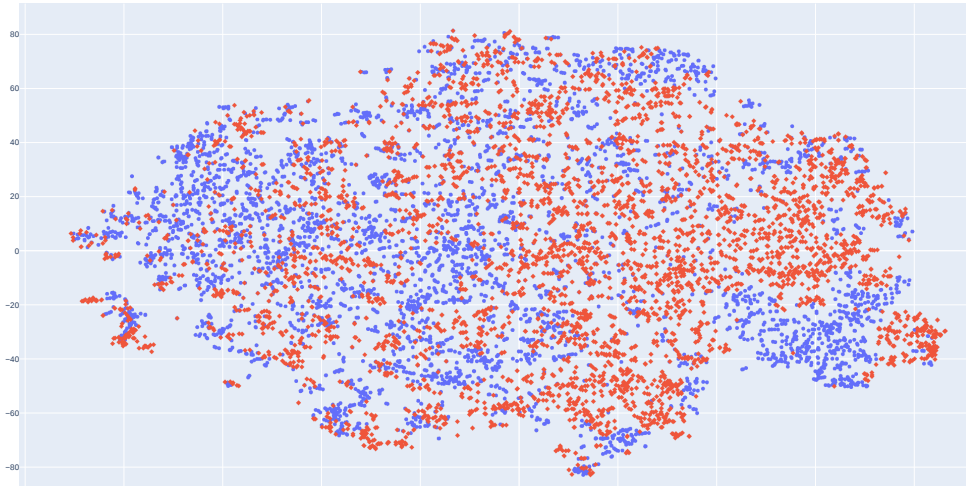

(b) With Romanized transcriptions (CKA Score=0.2024)

Figure 4: Repesentation visualization of semantically-aligned bilingual dictionaries in the case without Romanization (left) and with Romanization (right) between ZH (source) and VI (target) languages. Representations of source and target language are more aligned when Romanized transcriptions are introduced beyond the orthographic scripts. It is best to be viewed in colors.

cross-lingual settings, we replace the translated input with generated code-switched input from multilingual dictionaries; unlike machine translations, this code-switched input can be efficiently generated. We adopt the remaining training details, including hyperparameters, training schedules and evaluation metrics for downstream tasks, similarly to (Hu et al., 2020). We compare our framework to the following approaches:

• **XLM-R** : Fine-tuning the backbone *XLM-R-base* on source language ZH training data.
• **CoSDA-ML** (Qin et al., 2021): Multi-level code-switching augmentation for cross-lingual transfer.
• **FILTER** (Fang et al., 2021): Cross-lingual transfer via intermediate architecture disentanglement, with knowledge distillation objectives from source to target languages.
• **xTune** (Zheng et al., 2021): Two-stage augmentation mechanisms with four exhaustive augmentation methods for cross-lingual transfer.
• **X-MIXUP** (Yang et al., 2022): Cross-lingual transfer via manifold mixup augmentation and machine translation alignment between source and target languages.
• **PhoneXL** (Nguyen et al., 2023c): Cross-lingual transfer via additional IPA pre-training unsupervised alignment objectives to enhance token-level (NER and POS) task performance.

Due to space limitations, we present the concise average performance of JKV languages across multiple NLU tasks in Table 5. For individual language's performance for Sentence-level, Token-level and Question Answering tasks, we refer readers to Table 8, 9, 10 respectively in the Appendix.

## 8. Results

**Impact of Pre-segmentation Processing** As observed in Table 5, our proposed pre-processing provides tremendous benefits in target JKV languages' performance across multiple downstream tasks. For instance, fine-tuning XLM-R on our CORI benchmark, instead of raw XTREME, results in performance gain of 11.21% accuracy on Sentence-level XNLI tasks, 10.40 F1 points on Token-level PANX tasks and 4.08 points of EM on MLQA task. The gain is mainly due to the pre-segmentation processing as the backbone XLM-R fine-tuning only leverages orthographic scripts.

**Impact of Romanization Integration** Figure 4 demonstrates the effectiveness of Romanized transcription integration in which the representation discrepancy of 5000 randomly sampled semantically-aligned word entries in the bilingual dictionary between ZH and VI is reduced as compared to the orthographic-based representations.

Table 6: Ablation study on the impact of Romanization on representative multi-level NLU tasks.

| Source | PAWSX<br>Acc | PANX<br>F1 | XQuAD<br>EM |
|---|---|---|---|
| XLM-R (only Ortho) | 79.61 ± 5.24 | 37.29 ± 9.18 | 33.95 ± 14.73 |
| XLM-R (only Roman) | 60.23 ± 2.05 | 11.22 ± 3.17 | 22.77 ± 8.04 |
| **Ours (both)** | **81.38 ± 4.48** | **44.01 ±12.96** | **39.77 ± 10.50** |
| -CL | 80.07 ± 5.75 | 40.02 ± 15.64 | 35.21 ± 15.89 |

Additionally, we observe the consistent improvements in our orthographic-phonemic integration framework (*Ours(both)*) when compared to other variants in Table 6. This implies that the enhanced representation produced from our framework contributes to the improvements on zero-shot transfer performance across downstream tasks for JKV target languages. Despite the positive impact of Romanization integration, we observe that naive aggregation of Romanized transcription with orthographic counterpart without additional learning objectives is insufficient to enhance cross-lingual representation effectively. More importantly, as there exists a significant performance gap between *XLM-R (only Ortho)* and *XLM-R (only Roman)*, phonemic representation is not a direct substitute for orthographic counterpart, but a complimentary signal instead. We hypothesize the empirical results might be due to the impact of extensive exposure to orthographic textual scripts of PLMs during pretraining, leading to substantial gaps when training with only orthographic representation as compared to only Romanized transcription.

## 9. Conclusion

In our work, we conduct a preliminary study on the effect of source language selection in cross-lingual transfer for target languages, demonstrating the necessity of careful source language selection. The study prompts our effort to construct CORI, a CJKV Benchmark dataset, which not only covers diverse NLU tasks for CJKV languages but also captures the shared properties of word-level semantics and phonemic-orthographic language contact. We further conduct empirical studies and design a simple framework to demonstrate the essence of these properties in enhancing cross-lingual representation, leading to competitive zero-shot cross-lingual transfer performance across different downstream NLU tasks.

## 10. Limitations

As an attempt to provide an automatically generated CJKV dataset across various downstream tasks, we rely on the heuristics and translation quality from easily accessible off-the-shelf MT tools. Despite imperfections, the generated translations shed lights on our pilot study on the importance of source language selection and interconnections of CJKV languages beyond orthographic textual scripts.

Secondly, as our dataset is constructed from XTREME, CORI does not contain domain-specific datasets for CJKV languages such as dialogue domains (Xia et al., 2020; Nguyen et al., 2020, 2023a). By open-source our processing scripts, we seek to enable the potentials of generating domain-specific CJKV datasets in future works.

Lastly, our phonemic-orthographic aggregation remains simplistic and only achieves competitive performance on downstream tasks when compared to other cross-lingual methods. However, it is sufficient to demonstrate the contribution of integrating Romanized transcriptions with orthographic textual scripts for enhanced cross-lingual representations. As each language representation might benefit differently from each modality, we leave explorations of more dynamic aggregation mechanisms such as Mixture-of-Experts (Shazeer et al., 2017) for future work, which can result in the new state-of-the-art for cross-lingual transfer performance on target JKV languages.

## 11. Acknowledgement

This work is supported in part by NSF under grant III-2106758.

Table 7: Detailed preliminary study on the impact of source language on individual target JKV languages on individual tasks. **Bold** denote the **average performance (Avg)** together with **standard deviation (Std)** across 3 target JKV languages reported in Table 2.

| Source | Target | PAWSX | XNLI | UDPOS | PANX | XQuAD | MLQA |
|---|---|---|---|---|---|---|---|
| EN | JA | 70.70 | 69.72 | 28.70 | 17.98 | 13.86 | 14.35 |
| | KO | 68.53 | 69.13 | 42.76 | 20.77 | 22.77 | 20.37 |
| | VI | 81.35 | 66.19 | 48.52 | 52.34 | 48.52 | 31.61 |
| | **Avg** | **73.53** | **68.35** | **40.09** | **30.36** | **28.38** | **22.11** |
| | **Std** | **6.86** | **2.05** | **10.32** | **19.09** | **18.00** | **8.76** |
| ZH | JA | 76.98 | 73.18 | 46.24 | 47.50 | 23.92 | 24.12 |
| | KO | 76.20 | 71.02 | 47.19 | 29.70 | 27.07 | 29.84 |
| | VI | 85.65 | 66.93 | 50.04 | 34.67 | 50.87 | 35.89 |
| | **Avg** | **79.61** | **70.39** | **47.82** | **37.29** | **33.95** | **29.95** |
| | **Std** | **5.24** | **1.60** | **1.98** | **9.18** | **14.73** | **5.88** |

Table 8: Individual Language Performance results on **Sentence-level task** (XNLI and PAWSX). **Bold** denotes the average performance with standard deviation across 3 target languages reported in Table 5.

| Model | PAWSX | | | | XNLI | | | |
|---|---|---|---|---|---|---|---|---|
| | JA | KO | VI | Avg | JA | KO | VI | Avg |
| XLM-R (raw XTREME) | 64.82 | 77.75 | 85.81 | **76.13** ± **10.59** | 36.47 | 67.55 | 73.50 | **59.18** ± **19.89** |
| XLM-R | 76.98 | 76.20 | 85.65 | **79.61** ± **5.24** | 73.18 | 71.02 | 66.93 | **70.39** ± **1.60** |
| **Ours** | 78.35 | 79.10 | 86.45 | **81.38** ± **4.48** | 74.05 | 71.36 | 68.95 | **71.45** ± **2.55** |
| CoSDA-ML | – | – | – | – | – | – | – | – |
| FILTER | 76.85 | 75.82 | 86.30 | **79.66** ± **5.78** | 69.22 | 68.25 | 66.72 | **68.06** ± **1.09** |
| xTune | 78.65 | 79.12 | 87.38 | **81.72** ± **4.91** | 73.67 | 71.09 | 71.98 | **72.25** ± **1.31** |
| X-MIXUP | 76.98 | 76.43 | 87.81 | **80.41** ± **6.43** | 73.71 | 71.99 | 67.91 | **71.20** ± **2.98** |
| PhoneXL | – | – | – | – | – | – | – | – |

Table 9: Individual Language Performance results on **Token-level task** (UDPOS and PANX). **Bold** denotes the average performance with standard deviation across 3 target languages reported in Table 5.

| Model | UDPOS | | | | PANX | | | |
|---|---|---|---|---|---|---|---|---|
| | JA | KO | VI | Avg | JA | KO | VI | Avg |
| XLM-R (raw XTREME) | – | – | – | – | 34.57 | 15.20 | 30.88 | **26.89** ± **10.29** |
| XLM-R | 46.24 | 47.19 | 50.04 | **47.82** ± **1.98** | 47.50 | 29.70 | 34.67 | **37.29** ± **9.18** |
| **Ours** | 50.77 | 51.01 | 53.47 | **51.75** ± **1.49** | 58.70 | 39.12 | 34.21 | **44.01** ± **12.96** |
| CoSDA-ML | 46.33 | 52.38 | 52.73 | **50.48** ±**3.60** | 40.76 | 14.10 | 34.35 | **29.73** ± **13.91** |
| FILTER | 22.50 | 47.36 | 56.95 | **42.27** ± **17.78** | 28.33 | 25.31 | 34.38 | **29.34** ± **4.62** |
| xTune | 49.42 | 51.53 | 51.29 | **50.75** ± **1.16** | 60.70 | 25.54 | 35.34 | **40.53** ± **18.17** |
| X-MIXUP | 49.89 | 51.27 | 51.35 | **50.84** ± **0.82** | 44.62 | 37.47 | 40.06 | **40.71** ± **3.62** |
| PhoneXL | 47.76 | 49.30 | 51.30 | **49.45** ± **1.77** | 53.85 | 19.83 | 41.10 | **38.26** ± **17.19** |

Table 10: Individual Language Performance results on **Question Answering task** (XQuAD and MLQA). **Bold** denotes the average performance with standard deviation across 3 target languages reported in Table 5.

| Model | XQuAD | | | | MLQA | | | |
|---|---|---|---|---|---|---|---|---|
| | JA | KO | VI | Avg | JA | KO | VI | Avg |
| XLM-R (raw XTREME) | 17.60 | 23.48 | 49.86 | **30.31** ± **17.18** | 12.86 | 23.77 | 40.99 | **25.87** ± **14.18** |
| XLM-R | 23.92 | 27.07 | 50.87 | **33.95** ± **14.73** | 24.12 | 29.84 | 35.89 | **29.95** ± **5.88** |
| **Ours** | 31.92 | 35.68 | 51.70 | **39.77** ± **10.50** | 35.45 | 30.11 | 41.94 | **35.83** ± **5.92** |
| CoSDA-ML | – | – | – | – | – | – | – | – |
| FILTER | 31.09 | 22.52 | 42.72 | **32.10** ± **10.14** | 34.01 | 29.66 | 20.52 | **28.06** ± **6.89** |
| xTune | 31.02 | 41.68 | 53.50 | **42.07** ± **11.25** | 40.82 | 21.75 | 36.82 | **33.13** ± **10.06** |
| X-MIXUP | 30.67 | 16.69 | 50.72 | **32.69** ± **17.10** | 24.23 | 23.64 | 36.85 | **28.24** ± **7.46** |
| PhoneXL | – | – | – | – | – | – | – | – |